\documentclass[sigconf]{acmart}
\makeatletter\@ACM@balancefalse\makeatother

\usepackage{booktabs}
\usepackage{multirow}
\usepackage{graphicx}
\usepackage{amsmath}
\usepackage{enumitem}
\usepackage{tikz}
\usetikzlibrary{positioning,arrows.meta,backgrounds,calc}

\AtBeginDocument{}

\setcopyright{none}
\makeatletter\def\ACM@cc@type{}\makeatother
\renewcommand\footnotetextcopyrightpermission[1]{}
\copyrightyear{2026}
\acmYear{2026}
\acmDOI{}
\acmConference[Arxiv preprint]{}{2026}{}

\begin{document}

\title{MetaSyn: A Benchmark for LLM Agents on Meta-Analysis Articles from Nature Portfolio}
\thanks{Code: \url{https://github.com/THUIR/MetaSyn}; Dataset: \url{https://huggingface.co/datasets/THUIR/MetaSyn}.}

\author{Anzhe Xie}
\email{xaz25@mails.tsinghua.edu.cn}
\affiliation{\institution{Tsinghua University}\city{Beijing}\country{China}}

\author{Weihang Su}
\email{swh22@mails.tsinghua.edu.cn}
\affiliation{\institution{Tsinghua University}\city{Beijing}\country{China}}

\author{Yujia Zhou}
\email{zhouyujia@mail.tsinghua.edu.cn}
\affiliation{\institution{Tsinghua University}\city{Beijing}\country{China}}

\author{Yiqun Liu}
\email{yiqunliu@tsinghua.edu.cn}
\affiliation{\institution{Tsinghua University}\city{Beijing}\country{China}}

\author{Min Zhang}
\email{z-m@tsinghua.edu.cn}
\affiliation{\institution{Tsinghua University}\city{Beijing}\country{China}}

\author{Qingyao Ai}
\authornote{Corresponding author.}
\email{aiqy@tsinghua.edu.cn}
\affiliation{\institution{Tsinghua University}\city{Beijing}\country{China}}

\begin{abstract}
Systematic review and meta-analysis is an important method for scientific research. It comprehensively studies target research questions by combining evidence from multiple independent studies following a standard and rigid protocol (i.e., PI/ECO). This is valuable for facilitating reliable and sustainable research across multiple domains, including but not limited to physics, chemistry, psychology, and medical science. Traditional meta-analysis is both mind-intensive and labor-intensive, as it requires professionals to search, screen, and synthesize tens of thousands of studies, which usually takes months or even years. Recent advances in AI, particularly LLM agents, have the potential to significantly automate this process, but whether they can conduct reliable meta-analysis is still unknown. To this end, we introduce MetaSyn, a dataset and evaluation protocol built from 422 expert-curated meta-analyses manually selected from more than 34,000 published articles in Nature Portfolio journals. We collected the research questions with structured eligibility criteria, the studies included by the original reviewers, and a shared PubMed-anchored corpus containing both eligible studies and plausible but ineligible distractors to build a comprehensive benchmark for meta-analysis. Our experiments on multiple LLMs and baseline methods show that existing AI systems are far from perfect. We also conducted stage-wise evaluation and analysis to shed light on why existing AI systems fall short on meta-analysis.
\end{abstract}




\maketitle

\section{Introduction}
\label{sec:intro}

Scientific literature is growing faster than any individual reader can inspect it. In medicine, for example, dozens of new trials and systematic reviews appear each day, and unreliable synthesis can lead to significant resource waste and delayed translation into practice~\cite{bastian2010seventyfive,chalmers2009avoidable,elliott2014living}. Therefore, systematic review and meta-analysis, a protocol-driven way to combine results from multiple independent studies that address the same research question (for simplicity, we refer to it as \textit{meta-analysis}), is of great importance for scientific studies~\cite{glass1976primary,gurevitch2018meta}. In a typical meta-analysis, reviewers pre-specify where to search, which studies are eligible, what measurements to extract, and how those measurements will be aggregated~\cite{cochrane2019handbook,page2021prisma}. The way to evaluate whether a study should be included in meta-analysis, i.e., eligibility, is often described with PI/ECO criteria: Population, Intervention or Exposure, Comparison, and Outcome~\cite{schardt2007pico,morgan2018peco}. In contrast to a naive search-based summarization process (e.g., deep research provided by LLM services~\cite{gptresearcher,opendeepresearch}) that mostly retrieves documents by topical relevance or text quality, PI/ECO explicitly defines what kind of study counts as evidence for the target question following a scientific methodology, which makes the whole process verifiable and reproducible. Thus, final reports generated from the included studies are considered valuable and are often published in major scientific journals such as those in Nature Portfolio.

Despite its significance, the process of meta-analysis itself is also time-consuming and labor-intensive. While the rise of online scientific databases (e.g., PubMed) and scientific search engines (e.g., Google Scholar) improves the accessibility of scientific literature, retrieval, screening, extraction, and report writing based on PI/ECO are, as of today, still mostly performed manually by professionals~\cite{tsafnat2014automation,omaraeves2015textmining,marshall2019automation}. One of the primary reasons is that reading and filtering scientific literature on a specific question, while easier than actually doing the research, still requires a broad understanding of the field and strong academic skills. Traditional AI systems built with ad hoc task training can hardly handle such tasks, as they do not have the ability to generalize to unseen documents or studies. However, recent advances in large language models (LLMs) bring us brand new opportunities for meta-analysis. Existing studies have shown that LLMs built from large-scale training on both open-domain and domain-specific corpora can obtain human-like natural language understanding and processing abilities, and perform strongly in professional domains such as mathematics, law, and clinical work~\cite{deepseekr1nature,gpt5systemcard,Khraisha_2024}. This inspires us to ask the following question:

\begin{quote}
\textit{Can LLM agents conduct high-quality meta-analysis like human experts?}
\end{quote}

To answer this question, we introduce MetaSyn, a benchmark dataset to evaluate whether AI systems can conduct rigid meta-analysis in science. MetaSyn contains 422 meta-analysis articles manually selected from more than 34,000 candidate papers published in Nature Portfolio journals. We recruited trained human annotators to record and verify all the PI/ECO questions, search strategies, date bounds, and inclusion and exclusion criteria used in each meta-analysis, and to transcribe 15,921 entries from their published included-study lists. To better simulate the task scenarios of professional meta-analysis reviewers, we queried PubMed with the included-study titles to collect related articles and create a corpus with 140,585 articles. AI systems need to conduct retrieval, screening, and synthesis directly on this corpus to generate a meta-analysis report. Evaluation is done by comparing the AI-generated reports with the actual meta-analysis articles published in Nature Portfolio journals from multiple perspectives, including retrieval quality, screening accuracy, criteria adherence, and synthesis quality. In the experiments, we tested multiple LLM backbones with multiple methods, including the newly proposed ProtoMA method built around the standard meta-analysis pipeline. Results show that AI systems still fall short on multiple key parts of the meta-analysis process. Meta-analysis is also fundamentally different from traditional retrieval tasks. For example, high recall in first-stage retrieval does not guarantee better performance in the screening process, and the inclusion criteria generated by AI can differ significantly from those written by human experts. More research and better agentic frameworks are needed for LLMs to replace humans in scientific meta-analysis.

Our contributions are:

\begin{enumerate}[leftmargin=*]
\item \textbf{A large scientific meta-analysis dataset.} We provide a benchmark dataset with 422 expert-curated meta-analyses manually selected from Nature Portfolio. Each instance contains the corresponding review protocol, published included-study list, conclusion, and key findings annotated by human annotators, with stable article IDs linking the included articles to a shared PubMed corpus.
\item \textbf{An evaluation protocol.} We define stage-wise evaluation for retrieval, screening, and synthesis. The protocol separates where a system fails instead of reducing the full workflow to one end-to-end score.
\item \textbf{Experiments on multiple baseline systems.} We benchmark representative retrieval, RAG, and protocol-driven pipelines. The results show that current systems are far from perfect for meta-analysis.
\end{enumerate}

\section{Preliminaries}
\label{sec:pre}

Each meta-analysis follows a protocol-driven evidence workflow in which studies must satisfy predefined eligibility criteria to count as evidence. We call the expert-authored publication used to construct a benchmark instance the \textit{source review}.

\subsection{Terminology}
\label{sec:terminology}

Evidence synthesis involves several objects that are easy to confuse. Table~\ref{tab:core_terms} defines the terms used throughout the paper; Appendix~\ref{app:terminology} gives fuller definitions and examples. A \textit{study} is an investigation conducted to answer a research question, such as a clinical trial or an observational study. An \textit{article} is a publication that describes the methods or results of a study. A study may have multiple articles.

\begin{table}[H]
\centering
\caption{Core terminology. Article labels are defined separately for each benchmark instance.}
\label{tab:core_terms}
\small
\setlength{\tabcolsep}{3.5pt}
\begin{tabular}{@{}p{0.32\columnwidth}p{0.60\columnwidth}@{}}
\toprule
\textbf{Term} & \textbf{Meaning in MetaSyn} \\
\midrule
Source review & The published article used to build one benchmark instance and provide its protocol, included-study list, and synthesis. \\
Meta-analysis protocol & The PI/ECO eligibility rules and other stated constraints. \\
Included-study list & The studies or study reports that the source review says it included. \\
Evidence synthesis & The conclusion and key findings drawn from a selected body of evidence. \\
Primary study & A distinct investigation whose results are used as evidence; one study may be reported in several articles. \\
Corpus article & A PubMed article record in the shared MetaSyn corpus. \\
Included article & A corpus article matched by title to an entry in the source review's included-study list. \\
Reference set & All included articles linked to one benchmark instance and used to score it. \\
Candidate article & A corpus article considered for possible inclusion. \\
Predicted included article & An article explicitly listed or cited as included in the generated report. \\
\bottomrule
\end{tabular}
\end{table}

\subsection{Workflow and PI/ECO Criteria}

A meta-analysis usually proceeds in four ordered stages~\cite{cochrane2019handbook}:
\begin{enumerate}[leftmargin=*,itemsep=1pt,topsep=2pt]
    \item \textbf{Retrieval.} Retrieve candidate articles from academic databases.
    \item \textbf{Screen.} Apply the protocol to decide which studies are eligible.
    \item \textbf{Extract and analyze.} Extract comparable outcomes and combine them statistically when appropriate.
    \item \textbf{Synthesize and report.} Interpret the evidence and state the main conclusion, findings, and uncertainty.
\end{enumerate}
PRISMA provides a corresponding framework for reporting the completed meta-analysis~\cite{page2021prisma}. MetaSyn evaluates the three stages for which consistent reference labels can be constructed across the source reviews: retrieval, protocol-based selection, and written synthesis.

MetaSyn combines PICO~\cite{schardt2007pico} and PECO~\cite{morgan2018peco} in a PI/ECO representation. Its fields are Population, Intervention or Exposure, Comparison, and Outcome. We use them to structure each instance's eligibility rules. Two articles can look similar by topic but require different decisions because one uses the wrong population, comparator, follow-up window, or study design. Topical retrieval must therefore be followed by protocol-based screening.

\subsection{An Illustrative Example}
\label{sec:example}

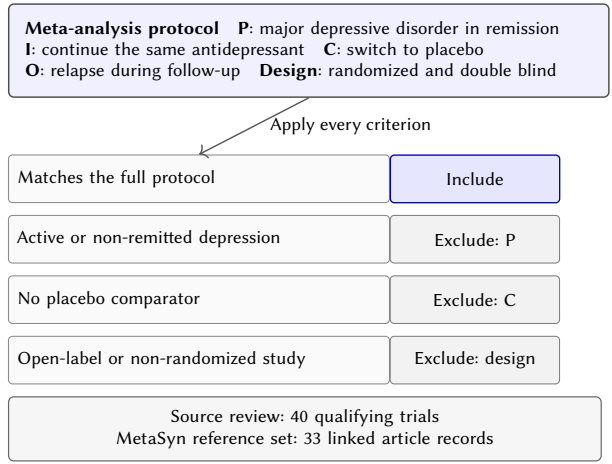
\begin{figure}[H]
\centering
\begin{tikzpicture}[
  font=\footnotesize\sffamily,
  protocol/.style={rectangle, rounded corners=2pt, draw=black!70, line width=0.55pt, fill=blue!6, text width=75mm, inner sep=2.2mm, align=left},
  candidate/.style={rectangle, rounded corners=1.5pt, draw=black!45, line width=0.4pt, fill=gray!4, text width=48mm, minimum height=6.3mm, inner sep=1.2mm, align=left},
  include/.style={rectangle, rounded corners=1.5pt, draw=blue!65!black, line width=0.5pt, fill=blue!10, text width=20mm, minimum height=6.3mm, inner sep=1.2mm, align=center},
  exclude/.style={rectangle, rounded corners=1.5pt, draw=black!50, line width=0.4pt, fill=gray!10, text width=20mm, minimum height=6.3mm, inner sep=1.2mm, align=center},
  summary/.style={rectangle, rounded corners=2pt, draw=black!60, line width=0.45pt, fill=gray!7, text width=75mm, inner sep=1.7mm, align=center},
  arrow/.style={->, line width=0.55pt, black!65}
]
\node[protocol, anchor=north west] (protocol) at (0mm,0mm) {\textbf{Meta-analysis protocol}\quad \textbf{P}: major depressive disorder in remission\\
\textbf{I}: continue the same antidepressant\quad \textbf{C}: switch to placebo\\
\textbf{O}: relapse during follow-up\quad \textbf{Design}: randomized and double blind};
\node[candidate, anchor=north west] (c1) at (0mm,-20mm) {Matches the full protocol};
\node[include, anchor=north west] (d1) at (50.5mm,-20mm) {Include};
\node[candidate, anchor=north west] (c2) at (0mm,-28mm) {Active or non-remitted depression};
\node[exclude, anchor=north west] (d2) at (50.5mm,-28mm) {Exclude: P};
\node[candidate, anchor=north west] (c3) at (0mm,-36mm) {No placebo comparator};
\node[exclude, anchor=north west] (d3) at (50.5mm,-36mm) {Exclude: C};
\node[candidate, anchor=north west] (c4) at (0mm,-44mm) {Open-label or non-randomized study};
\node[exclude, anchor=north west] (d4) at (50.5mm,-44mm) {Exclude: design};
\draw[arrow] (protocol.south) -- node[right, xshift=1mm, font=\footnotesize\sffamily, text=black] {Apply every criterion} (c1.north);
\node[summary, anchor=north west] (summary) at (0mm,-52mm) {Source review: 40 qualifying trials\\MetaSyn reference set: 33 linked article records};
\end{tikzpicture}
\caption{Conceptual protocol-based screening for one MetaSyn instance. Topic-related candidates are eligible for inclusion only when they satisfy every eligibility rule. Trial and article counts use different units.}
\Description{A meta-analysis protocol specifies population, intervention, comparison, outcome, and study design. Four example candidates pass through the protocol. A matching candidate is included, while candidates with active depression, no placebo comparator, or an open-label design are excluded. The source review reports 40 qualifying trials, and MetaSyn links 33 article records from its published included-study list.}
\label{fig:protocol_example}
\end{figure}

Consider the research question \textit{Does continuing antidepressant treatment after remission prevent relapse in patients with major depressive disorder}~\cite{kato2021discontinuation}? Figure~\ref{fig:protocol_example} shows how its protocol separates eligible evidence from closely related candidates.

\section{Dataset Construction}
\label{sec:data}

MetaSyn turns each published source review into one benchmark instance whose labels can be checked against the publication. The source review provides an inspectable protocol, states which studies or articles its authors included, and reports its synthesis of that evidence. Most source reviews report quantitative results. Creating a common table of study-level effects requires separate extraction and harmonization from the linked articles. The PubMed IDs provide stable starting points for this work. MetaSyn currently evaluates retrieval, selection, and the written synthesis.

Construction proceeds in four stages: search the \textit{Nature} Portfolio, screen open-access articles for an inspectable included-study list, annotate and verify the protocol and synthesis fields, and match the listed titles to PubMed records. The final dataset contains 422 meta-analysis instances whose included-study lists can be inspected in the source review or its linked supplementary materials. We place the matched article records and related candidates in a shared corpus of 140,585 articles. The benchmark asks systems to find the articles selected by the source review and synthesize their evidence.

\subsection{Source Review Collection and Selection}
\label{sec:source}

We drew source reviews from journals in the \textit{Nature} Portfolio because their protocols, supplements, and included-study lists are often available in formats that annotators can inspect consistently. The portfolio spans clinical, biological, social-science, environmental, and digital-health topics. This choice makes annotation manageable while retaining varied research questions and eligibility rules.

We searched for articles containing ``meta-analysis'' or ``systematic review'' across the \textit{Nature} Portfolio, yielding over 60,000 keyword hits. After deduplication and scope filtering, 34,375 candidates remained for manual screening.

\subsection{Human Annotation and Filtering}

Approximately 50 annotators participated in dataset collection and verification. They received the same written instructions, task-specific guidelines, and worked examples. They screened all 34,375 candidates and kept open-access articles whose complete included-study list could be read from the article, its supplements, linked data files, or the references shown in forest plots. For retained instances, annotators recorded the research question, search information, eligibility criteria, included studies, and synthesis fields. After unreliable extractions and records with more than 300 included entries were removed, 422 meta-analysis instances remained.

To ensure annotation quality, every retained source review underwent a second verification pass against the article and its supplements. This pass checked the included-study list and annotated fields. PubMed matching was then performed from the verified titles. Uncertain records were corrected or removed. Appendix~\ref{app:annotation_provenance} documents this process.

\subsection{Protocol Structuring and Synthesis Labels}
\label{sec:rq}

Research questions vary across source reviews, and eligibility rules often appear in several sections. We therefore put each question into a consistent PI/ECO format. A small set was structured manually; GLM-4.6~\cite{zai2025glm46} drafted most remaining PI/ECO fields from the source review alone, and annotators corrected and verified them. The Intervention or Exposure field records the intervention in intervention studies and the exposure in observational studies. We leave Comparison empty when the source review does not state one. We also extract synthesis labels from the source review itself. Annotators record its main conclusion, key findings, and one of three conclusion-direction labels: Positive, Negative, or Mixed. Positive and Negative indicate the direction of the reported answer. Mixed covers conflicting, null, heterogeneous, descriptive, or otherwise non-directional conclusions. These labels provide a consistent reference for judging whether a system preserves the source review's conclusion and key findings.

\subsection{Corpus Construction}

To create a realistic retrieval setting, the corpus must contain both reference articles and plausible alternatives. While screening candidate meta-analyses, we searched PubMed for titles from their included-study lists and retained the matched records and related search results. We also retained a smaller set of randomly sampled PubMed records from the first-pass pool. After deduplication and consolidation, this process produced a shared corpus of 140,585 articles.

For the final 422 meta-analysis instances, title matching produces one reference set per instance. The same article can belong to more than one reference set; Table~\ref{tab:data_stats} reports the number of distinct linked articles. All other corpus articles remain candidates for that instance.

MetaSyn retains every transcribed included article title, including titles without a PubMed match. Linked PubMed IDs define the current reference sets, and the full title lists support future matching through other literature databases.

Each corpus record retains its PubMed ID as a stable pointer to the source record. When PubMed links an article to full text in PubMed Central (PMC), we store its structured sections. In total, 80,158 articles (57.0\%) have PMC full text, and 67,961 articles (48.3\%) contain at least one section beyond the abstract, such as introduction, methods, results, or conclusion. The corpus therefore supports experiments with abstracts alone or with the available full text.

\subsection{Dataset Statistics}
\label{sec:stats}

Table~\ref{tab:data_stats} summarizes MetaSyn. MetaSyn contains 422 instances, each with an instance-specific reference set of PubMed-linked articles. An instance/article pair links one instance to one article in its reference set; the same PubMed article can contribute pairs to several instances. The dataset contains 7,374 such pairs, covering 7,187 unique article IDs. The test split contains 1,677 pairs and 1,649 unique IDs. Per instance, the full dataset averages 38.4 reported included studies or articles and 17.5 linked articles; the test set averages 29.3 and 19.5. For each instance, we divide linked articles by the number of nonempty titles transcribed from its published included-study list, then macro-average the instance-level rates. The resulting title-match rate is 51.6\% overall and 67.7\% on the test set.

Appendix~\ref{app:dataset_coverage} reports domain coverage and additional fields, while Appendix~\ref{app:release_scope} describes release and versioning.

\begin{table}[t]
\centering
\caption{MetaSyn dataset statistics. All evaluations use the held-out test split.}
\label{tab:data_stats}
\small
\setlength{\tabcolsep}{4pt}
\begin{tabular}{@{}lcc@{}}
\toprule
\textbf{Statistic} & \textbf{Full} & \textbf{Test} \\
\midrule
Meta-analysis instances & 422 & 86 \\
Avg. included studies/articles reported & 38.4 & 29.3 \\
Instance/article pairs & 7,374 & 1,677 \\
Unique PubMed-linked articles & 7,187 & 1,649 \\
Avg. linked articles per instance & 17.5 & 19.5 \\
Title-match rate, macro avg. & 51.6\% & 67.7\% \\
PI/ECO structure available & 100\% & 100\% \\
Search strategies available & 99.5\% & 98.8\% \\
Search end date available & 99.5\% & 98.8\% \\
Inclusion criteria available & 96.4\% & 97.7\% \\
\midrule
Corpus articles & \multicolumn{2}{c}{140,585} \\
Official PMC full text & \multicolumn{2}{c}{80,158 (57.0\%)} \\
Structured full-text sections & \multicolumn{2}{c}{67,961 (48.3\%)} \\
\bottomrule
\end{tabular}
\vspace{0.25em}
\begin{flushleft}
\footnotesize ``Avg. included studies/articles reported'' is the count stated by the source review. An instance/article pair links one instance to one article; the unique linked-article count deduplicates PubMed IDs across instances. Title-match rate uses the nonempty titles transcribed from each included-study list and is macro-averaged across instances. PI/ECO uses Intervention or Exposure as applicable; Comparison is left empty when it is not stated.
\end{flushleft}
\end{table}

\section{Benchmark Design and Metrics}
\label{sec:benchmark}

\subsection{Tasks and Labels}
MetaSyn defines two linked tasks. The retrieval task ranks corpus articles for a meta-analysis instance. The end-to-end task uses the retrieved evidence to produce a report that names the included articles, states the inclusion and exclusion criteria, and gives the conclusion and key findings.

For each instance, the \textit{reference set} $G$ is the set of PubMed-linked corpus articles matched to an entry in its source review's published included-study list. This instance-specific set is the evidence target for scoring. An article ID can occur in several instance-specific reference sets, which is why pair counts and unique article counts differ.

This release scores linked PubMed IDs and retains the full title lists for future database linkage.

We score the initial ranked list first, then the generated report after evidence selection and synthesis. Together, these evaluations attribute each missing reference article to retrieval or to later selection and reporting. We use \textit{selection} for the overall process of choosing evidence and \textit{screening} for a candidate-level include or exclude decision. The inclusion metrics use the articles explicitly listed or cited in the final report; ProtoMA also exposes candidate-level screening decisions.

\begin{figure*}[!t]
\centering
\begin{tikzpicture}[
  font=\sffamily,
  stage/.style  = {rectangle, rounded corners=3pt, draw=black!70, line width=0.65pt, fill=blue!8, text width=30mm, minimum height=17mm, inner sep=1.5pt, outer sep=0pt, align=center, font=\small\sffamily},
  gt/.style     = {rectangle, rounded corners=2pt, draw=black!55, line width=0.45pt, fill=green!8, text width=30mm, minimum height=17mm, inner sep=1.5pt, outer sep=0pt, align=center, font=\small\sffamily},
  metric/.style = {rectangle, draw=black!30, line width=0.4pt, fill=orange!8, text width=30mm, minimum height=17mm, inner sep=1.5pt, outer sep=0pt, align=center, font=\small\sffamily},
  io/.style     = {rectangle, rounded corners=4pt, draw=black!70, line width=0.65pt, fill=gray!10, text width=22mm, minimum height=17mm, inner sep=1.5pt, outer sep=0pt, align=center, font=\small\sffamily},
  arrow/.style  = {-{Latex[length=2mm]}, line width=0.7pt, black!60},
  gtarrow/.style= {-{Latex[length=1.8mm]}, line width=0.5pt, dashed, black!55},
  lbl/.style    = {text width=22mm, font=\small\itshape\sffamily, black!60, align=center}
]
\node[io] (input) {Research Question\\ + PI/ECO};
\node[stage, right=3.5mm of input] (retr) {\textbf{Retrieval}\\Query $\rightarrow$ Top-$K$\\140{,}585-Article Corpus};
\node[stage, right=3.5mm of retr]  (scr)  {\textbf{Evidence Selection}\\Screen Candidates\\And List Selected\\Articles};
\node[stage, right=3.5mm of scr]   (syn)  {\textbf{Report Synthesis}\\Direction\\+ Insights};
\node[io, right=3.5mm of syn] (out) {Structured\\Evidence-Synthesis\\Report};
\draw[arrow] (input) to (retr);
\draw[arrow] (retr)  to (scr);
\draw[arrow] (scr)   to (syn);
\draw[arrow] (syn)   to (out);
\node[gt, above=5.5mm of retr] (gtr)  {ID Labels:\\Included Articles};
\node[gt, above=5.5mm of scr]  (gts)  {ID + Protocol Labels:\\Included Articles,\\Eligibility Criteria};
\node[gt, above=5.5mm of syn]  (gty)  {Synthesis Labels:\\Direction\\+ Key Insights};
\node[lbl] (gtlabel) at (input |- gtr) {Reference\\Labels};
\draw[gtarrow] (gtr.south) to (retr.north);
\draw[gtarrow] (gts.south) to (scr.north);
\draw[gtarrow] (gty.south) to (syn.north);
\node[metric, below=5.5mm of retr] (mr) {R@$K$, P@$K$};
\node[metric, below=5.5mm of scr]  (ms) {Inc.R / Inc.P / Inc.F1\\Scr.A\\Criteria Consistency};
\node[metric, below=5.5mm of syn]  (my) {Direction Accuracy\\Insights\\Structure};
\node[lbl] (mlabel) at (input |- mr) {Evaluation\\Metrics};
\draw[gtarrow] (retr.south) to (mr.north);
\draw[gtarrow] (scr.south)  to (ms.north);
\draw[gtarrow] (syn.south)  to (my.north);
\begin{scope}[on background layer]
  \path[fill=black!3, rounded corners=4pt]
    ($(retr.north west)+(-1.5mm,2mm)$) rectangle ($(syn.south east)+(1.5mm,-2mm)$);
\end{scope}
\end{tikzpicture}
\caption{The evidence-synthesis workflow and MetaSyn's stage-wise evaluation. The middle row shows the workflow. The top row gives the reference labels extracted from the source review; the bottom row gives the metrics.}
\Description{Flow diagram from research question and PI/ECO input through retrieval, evidence selection, and report synthesis to a structured report, with reference labels and evaluation metrics attached to each stage.}
\label{fig:pipeline}
\end{figure*}
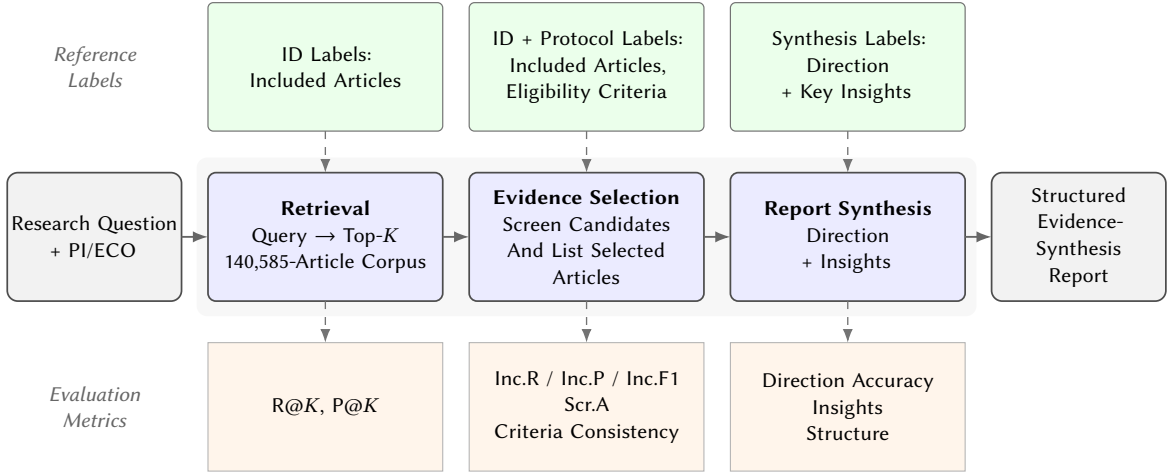

Figure~\ref{fig:pipeline} summarizes the standard workflow and the separate reference labels and metrics that MetaSyn adds at each stage.

\subsection{Retrieval Metrics}

The retrieval task receives the research question and structured PI/ECO fields and returns a ranked list from the shared corpus. Let $P_K$ denote the top-$K$ pool.

\begin{equation}
\mathrm{R@}K=\frac{|P_K\cap G|}{|G|},
\qquad
\mathrm{P@}K=\frac{|P_K\cap G|}{|P_K|}.
\label{eq:retrieval_metrics}
\end{equation}
R@$K$ is the share of reference articles retrieved, and P@$K$ is the share of the retrieved pool that belongs to the reference set. Both are computed per instance and then macro-averaged over the test set. They are reported separately from the end-to-end inclusion metrics. Token use is a retrieval-depth diagnostic described with its experiment.

\subsection{End-to-End Metrics}

A system receives the research question and PI/ECO elements. Baselines that use additional recorded protocol fields are identified in the experimental setup.

\paragraph{Included article metrics.}
The output names included articles by title or corpus ID; titles are matched against retrieved corpus articles. For one instance, $G$ is its set of linked reference article IDs and $P$ is the set of deduplicated candidate IDs exposed to the system. Let $L$ be the deduplicated final list after replacing each matched entry with its corpus ID; unmatched entries remain in $L$. Then
\begin{equation}
\begin{aligned}
\mathrm{Inc.R}&=\frac{|L\cap G|}{|G|}, &
\mathrm{Inc.P}&=\frac{|L\cap G|}{|L|},\\
\mathrm{Inc.F1}&=\frac{2\,\mathrm{Inc.R}\,\mathrm{Inc.P}}
 {\mathrm{Inc.R}+\mathrm{Inc.P}}.&&
\end{aligned}
\label{eq:id_metrics}
\end{equation}
Inc.R is the coverage of $G$, Inc.P is the fraction of $L$ that belongs to $G$, and Inc.F1 is their harmonic mean. Unmatched entries remain in the precision denominator and receive no credit. A zero denominator yields zero.

\paragraph{Screening accuracy.}
Let $\widehat{y}_a=\mathbf{1}[a\in L]$ be the predicted label and $y_a=\mathbf{1}[a\in G]$ the reference label for candidate $a$. Screening Accuracy is the fraction of exposed candidates with matching labels:
\begin{equation}
\mathrm{Scr.A}=\frac{1}{|P|}\sum_{a\in P}
 \mathbf{1}[\widehat{y}_a=y_a] .
\label{eq:screening_accuracy}
\end{equation}
Here $\mathbf{1}[\cdot]$ is $1$ when its condition holds and $0$ otherwise. This final-list rule is used consistently in the main table. ProtoMA's candidate-level records support the separate stage-loss analysis.

\paragraph{Criteria consistency.}
Let $Q$ be the criterion items extracted from the generated report, $R$ the reference criterion items, and $s(q,r)$ their embedding cosine similarity. We compute
\begin{equation}
\begin{gathered}
p_s=\frac{1}{|Q|}\sum_{q\in Q}\max_{r\in R}s(q,r),
\qquad
r_s=\frac{1}{|R|}\sum_{r\in R}\max_{q\in Q}s(q,r),\\[2pt]
\mathrm{Cons}(Q,R)=\frac{2p_sr_s}{p_s+r_s}.
\end{gathered}
\label{eq:criteria_metrics}
\end{equation}
Here $p_s$ is soft precision, the mean best-match similarity from each reported criterion to $R$. Likewise, $r_s$ is soft recall, the mean best-match similarity from each reference criterion to $Q$. $\mathrm{Cons}(Q,R)$ is their harmonic mean. Inc.C applies it to inclusion criteria and Exc.C to exclusion criteria. An empty $Q$ or $p_s+r_s\leq0$ scores zero; instances without $R$ are omitted.

\paragraph{Synthesis metrics.}
Let $d$ and $\widehat d$ be the reference and extracted conclusion directions. Then $\mathrm{Dir.A}=\mathbf{1}[\widehat d=d]$, which is $1$ for a matching label and $0$ otherwise. Insights measures coverage of the reference key findings from 0 to 1, and Structure Quality (SQ) rates report organization from 1 to 5. Table~\ref{tab:pipeline_results} reports all nine metrics. Scores are computed per instance and macro-averaged over instances with the required label; tables use percentages except for SQ. Appendix~\ref{app:evaluator_setup} gives the evaluator settings. Human validation in Appendix~\ref{app:validation_results} finds the strongest rank agreement for Dir.A ($\rho=+0.82$); Exc.C and Insights support system-level comparisons, while Inc.C and SQ are descriptive measures.

\section{Experiments}
\label{sec:experiments}

\subsection{Baselines}

We compare three retrievers. \textbf{BM25}~\cite{bm25} is the sparse baseline. \textbf{BGE-large-en-v1.5}~\cite{bge} is the general dense baseline with a FAISS index~\cite{faiss}. \textbf{Meta-Analysis Retriever (MA-Retriever)} fine-tunes BGE on pairs of structured queries and included articles from the MetaSyn training split.

One-pass RAG supplies a fixed top-200 list in rank order and asks an LLM to write the report from article titles, years, and abstracts. We test DeepSeek-V4-Pro, GLM-5.1~\cite{glm5}, and GPT-5.4~\cite{gpt5systemcard}. The prompt includes recorded dates and stated eligibility criteria, while retrieval uses the question and PI/ECO. The pipeline moves directly from the fixed ranked pool to report generation in one LLM call. We extract predicted included articles from the corpus IDs or article titles listed in the final report.

\textbf{Protocol Meta-Analysis (ProtoMA)} uses GPT-5.4 and adds explicit screening before synthesis. It generates three to five MA-Retriever queries, merges their results under a cap of 200, and screens the resulting titles and abstracts; the observed pool averages 72.3 articles. It then extracts information from retained articles and writes the report. We evaluate this complete workflow, including query generation, screening, extraction, and report construction.

\textbf{GPT-Researcher}~\cite{gptresearcher} and \textbf{Open Deep Research} (OpenDR)~\cite{opendeepresearch} retain their iterative workflows. Both use GPT-5.4 and MA-Retriever for planning, local retrieval, and report writing. Each retrieval call returns up to 20 records, with at most 200 distinct article IDs visible per run. GPT-Researcher and OpenDR expose averages of 47.0 and 36.1 distinct articles, respectively, and at most 82. Their prompts include the recorded dates and eligibility criteria. We compare retrieval variants within the same workflow family.

\subsection{Experimental Setup}

The main retrieval and end-to-end evaluations use all 86 held-out instances. Test-linked articles are excluded from MA-Retriever's positive training pairs. All three retrievers query with the research question and PI/ECO fields, and the source review is removed before top-$K$ truncation. Appendices~\ref{app:retrieval_setup} and~\ref{app:generation_setup} give the retriever and generation settings; Appendix~\ref{app:split_scope} describes the released split.

Faithfully reconstructing a source review should respect its search endpoint. A hard year filter is not reliable for the main comparison because the available dates do not identify database availability precisely: some source reviews lack a clear cutoff or report an updated search, and online publication, issue assignment, and indexing can occur in different years. We therefore use the same unfiltered corpus and evaluate recorded search end years separately.

\subsection{Retrieval-Only Results}

\begin{figure}[t]
\centering
\includegraphics[width=\columnwidth]{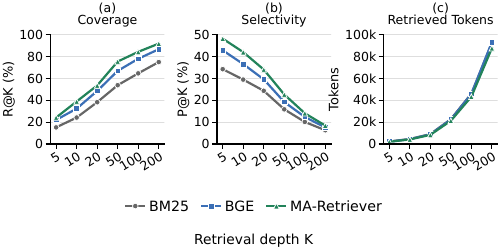}
\caption{Retrieval depth on 86 test instances. Panels (a) and (b) show recall and precision for BM25, BGE, and MA-Retriever. Panel (c) shows mean tokens in returned titles and abstracts, counted with \texttt{cl100k\_base}.}
\Description{Three line charts show retrieval recall, retrieval precision, and title-plus-abstract workload as retrieval depth increases.}
\label{fig:retrieval_tradeoff}
\end{figure}

We first measure retrieval without report generation. BM25, BGE, and MA-Retriever receive the same inputs and are evaluated at the same depths on all 86 test instances. Panels (a) and (b) of Figure~\ref{fig:retrieval_tradeoff} report the main R@$K$ and P@$K$ results.

\textbf{Task-specific retrieval improves coverage, while deeper pools become less selective.} At $K=100$, BM25, BGE, and MA-Retriever reach 64.6\%, 77.8\%, and 84.2\% recall. MA-Retriever rises to 91.7\% R@200, but its precision falls from 34.0\% at $K=20$ to 8.4\% at $K=200$. Its average top-200 pool contains about 17 reference articles and 183 other candidates.

To explain what remains after strong retrieval, we analyze MA-Retriever's high-ranked candidates that were not selected by their source reviews and its two hardest test instances. These diagnostics use title patterns to group publication types, followed by inspection of the recorded PI/ECO fields and candidate abstracts.

\textbf{High-ranked candidates still require protocol screening.} Among 1,135 top-20 records not selected by the corresponding source review, title patterns identify 167 (14.7\%) as other evidence syntheses and 38 (3.3\%) as protocols or other non-primary publications. The remaining 930 cases (81.9\%) generally require protocol fields, abstracts, or full text to determine eligibility. Appendix~\ref{app:retrieval_cases} gives concrete cases involving study design, comparator, outcome, and intervention details.

\textbf{The two hardest instances combine constraints that one PI/ECO query does not fully capture.} MA-Retriever finds fewer than 30\% of their reference articles in the top 100, and all three retrievers struggle on both. One asks how women participate in water, sanitation, and hygiene intervention delivery, a role seldom stated in article titles. The other combines causal evidence on salt intake from observational studies, Mendelian randomization studies, and randomized trials. Their central constraints concern participant roles and study design, which are weakly expressed in the structured query, title, and abstract.

\textbf{We test recorded temporal bounds separately.} Table~\ref{tab:date_filtering_main} compares the main and date-aware top-20 rankings; same-year and missing-year records remain.

\begin{table}[t]
\centering
\caption{Effect of date filtering on MA-Retriever top-20 retrieval. Date-aware removes records published after the recorded search end year and refills the pool.}
\label{tab:date_filtering_main}
\small
\renewcommand{\arraystretch}{0.96}
\setlength{\tabcolsep}{4pt}
\begin{tabular}{@{}llcc@{}}
\toprule
\textbf{Setting} & \textbf{Date Rule} & \textbf{R@20 (\%)} & \textbf{P@20 (\%)} \\
\midrule
Main & No year filter & 53.5 & 34.0 \\
Date-aware & Remove later years & 66.4 & 43.9 \\
\bottomrule
\end{tabular}
\end{table}

\textbf{Recorded time bounds improve retrieval but do not replace screening.} R@20 rises by 12.9 points and P@20 by 9.9 points. The result confirms that temporal bounds matter, while publication year remains an imperfect proxy for database availability. The filtered pool still requires protocol screening; Appendix~\ref{app:date_filtering} gives the full construction and candidate analysis.

\subsection{Retrieval Depth and Token Use}
\label{sec:token_use}

Retrieving more articles can improve coverage while increasing the evidence passed to later stages. Panel (c) of Figure~\ref{fig:retrieval_tradeoff} measures returned title-and-abstract tokens across all 86 instances. A 12-instance subset stratified by reference-set size measures GPT-5.4 token use for one-pass RAG and GPT-Researcher. Appendix~\ref{app:token_measurement} records the subset and accounting method.

\textbf{Deeper retrieval increases screening workload sharply.} Across all 86 instances, returned text grows from about 2.0k tokens at $K=5$ to 87.5k at $K=200$.

\textbf{For one-pass RAG, token use grows faster than final inclusion recall.} From $K=5$ to 200 on the subset, returned text grows from 1.9k to 88.0k tokens and total model use from 7.0k to 95.3k. Retrieval recall rises from 33.4\% to 91.0\%, while Inc.R rises from 17.5\% to 36.5\%; its 3.9-point decrease from $K=100$ to 200 is descriptive on this subset.

\textbf{Repeated retrieval calls increase token use.} Raising GPT-Researcher's $K$ from 5 to 20 grows returned text from 33.1k to 132.3k tokens and model use from 132.6k to 228.8k. Its 13 calls return 65 to 260 article instances but only 13.8 to 48.2 unique articles; retrieval recall rises from 50.2\% to 76.4\% and Inc.R from 26.0\% to 36.0\%. To separate call count from depth, we analyze all 86 OpenDR runs at fixed $K=20$. Total model tokens increase with retrieval-call count ($r=0.37$, $p<0.001$). Appendix~\ref{app:token_results} gives the full plots and accounting details.

\subsection{End-to-End Results}

We next evaluate the complete reports from every main configuration on all 86 test instances. Table~\ref{tab:pipeline_results} presents the main evidence-selection, criteria, and synthesis results. We first compare these main results, then use output and workflow traces to explain where their differences arise.

\begin{table*}[t]
\centering
\caption{End-to-end results on 86 test instances. Values are percentages except SQ (1 to 5); Inc.C uses 84 instances. Inclusion scores use matched corpus IDs and count unmatched citations as errors. One-pass RAG and the agents receive recorded criteria; ProtoMA generates them from PI/ECO. Bold marks column maxima.}
\label{tab:pipeline_results}
\footnotesize
\renewcommand{\arraystretch}{0.92}
\begin{tabular}{@{}llccccccccc@{}}
\toprule
 & & \multicolumn{3}{c}{\textbf{Evidence Set}} & \multicolumn{3}{c}{\textbf{Screening / Criteria}} & \multicolumn{3}{c}{\textbf{Report}} \\
\cmidrule(lr){3-5}\cmidrule(lr){6-8}\cmidrule(l){9-11}
\textbf{System} & \textbf{Retrieval} & \textbf{Inc.R} & \textbf{Inc.P} & \textbf{Inc.F1} & \textbf{Scr.A} & \textbf{Inc.C} & \textbf{Exc.C} & \textbf{Dir.A} & \textbf{Insights} & \textbf{SQ} \\
\midrule
\multirow{3}{*}{RAG DeepSeek-V4-Pro} & BM25 & 45.4 & 70.5 & 50.6 & 96.2 & \textbf{62.9} & 53.7 & 59.3 & 32.0 & 3.06 \\
 & BGE & 49.2 & 74.6 & 54.2 & 95.6 & 62.3 & 53.3 & 52.3 & 32.8 & 3.01 \\
 & MA-Retriever & \textbf{51.2} & 71.1 & 54.5 & 94.8 & 62.8 & 52.8 & 51.2 & 31.7 & 3.12 \\
\midrule
\multirow{3}{*}{RAG GLM-5.1} & BM25 & 41.5 & 71.1 & 48.6 & \textbf{96.4} & 57.7 & 44.0 & 55.8 & 32.7 & 2.48 \\
 & BGE & 48.5 & 76.7 & 54.7 & 95.8 & 56.0 & 46.0 & \textbf{61.6} & 33.7 & 2.56 \\
 & MA-Retriever & 51.0 & 73.5 & \textbf{56.0} & 95.1 & 59.0 & 45.2 & 51.2 & 32.1 & 2.52 \\
\midrule
\multirow{3}{*}{RAG GPT-5.4} & BM25 & 42.2 & 76.0 & 51.4 & 96.3 & 54.8 & 49.6 & 52.3 & 31.5 & 2.59 \\
 & BGE & 46.8 & \textbf{79.1} & 55.1 & 95.8 & 55.7 & 48.4 & 59.3 & 34.9 & 2.74 \\
 & MA-Retriever & 48.4 & 76.5 & 55.8 & 95.1 & 56.8 & 47.4 & 55.8 & 34.7 & 2.65 \\
\midrule
ProtoMA GPT-5.4 & MA-Retriever & 39.3 & 62.9 & 44.1 & 86.5 & 60.6 & 53.7 & 59.3 & 27.8 & 3.95 \\
\midrule
GPT-Researcher & MA-Retriever & 30.3 & 76.7 & 40.3 & 83.6 & 61.1 & 57.7 & 52.3 & \textbf{35.8} & \textbf{4.07} \\
\midrule
OpenDR & MA-Retriever & 27.2 & 75.6 & 36.3 & 80.5 & 59.0 & \textbf{58.2} & 51.2 & 27.0 & 3.92 \\
\bottomrule
\end{tabular}
\end{table*}

\begin{figure}[H]
\centering
\includegraphics[width=0.94\columnwidth]{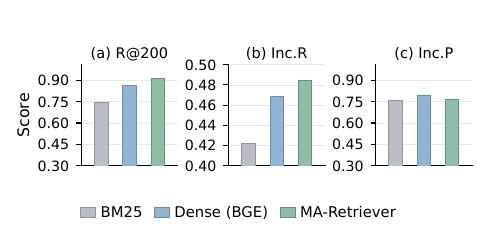}
\caption{GPT-5.4 one-pass RAG with three retrievers at $K=200$. One LLM call writes a report from the fixed 200-article pool, with no further retrieval. Pool R@200 is the share of reference articles retrieved; Inc.R and Inc.P are recall and precision for articles listed in the report. Bars are macro averages over 86 instances.}
\Description{Three metric groups compare retrieval recall, final inclusion recall, and final inclusion precision across BM25, BGE, and MA-Retriever.}
\label{fig:retrieval_to_rag}
\end{figure}

\textbf{Better retrieval raises final recall, but downstream losses remain.} With MA-Retriever, the top-200 pool covers 91.7\% of the reference set, while the three RAG final lists cover 48.4\% to 51.2\%. From BM25 to MA-Retriever, Inc.R rises by 5.8, 9.5, and 6.2 points for DeepSeek-V4-Pro, GLM-5.1, and GPT-5.4. Figure~\ref{fig:retrieval_to_rag} shows this gap for GPT-5.4.

\textbf{Final-list length explains part of the recall and precision tradeoff.} The fixed-pool systems list 10.2 to 12.6 articles per instance with MA-Retriever. GPT-Researcher and OpenDR list only 5.5 and 4.5, reaching 76.7\% and 75.6\% Inc.P but 30.3\% and 27.2\% Inc.R. The three RAG models list 8.5 to 9.5 reference articles on average, with only 0.01 to 0.13 unmatched entries. Their precision differences therefore arise mainly from selection among mapped corpus articles. Appendix~\ref{app:output_diagnostics} gives all list and citation statistics.

\textbf{Saved traces locate losses at retrieval and selection.} ProtoMA retrieves 79.3\% of the reference set, positively screens 49.0\% of those retrieved reference articles, and finally lists 39.3\% of the full reference set. Every positive screening ID appears in its final list, placing its measured downstream loss at explicit screening. Across all twelve standard configurations, retrieved-pool union recall is 95.0\%, while final-list union recall is 75.7\%. The 132 links missed by every pool show remaining room for retrieval, and the larger final-list gap shows additional loss after retrieval. Appendix~\ref{app:pipeline_traces} gives the full counts.

\textbf{Supplying criteria improves criteria agreement more than evidence selection.} The main comparison provides recorded criteria to hold the protocol fixed. We rerun GPT-5.4 RAG and GPT-Researcher without them while keeping all other settings fixed. The supplied text raises agreement with the reference criteria, most clearly for GPT-Researcher, whose Inc.C and Exc.C rise by about ten points. RAG Inc.R changes by only 1.0 point; GPT-Researcher shifts toward higher recall ($+5.5$) and lower precision ($-5.3$). Criteria availability alone therefore does not explain the evidence gap. The results point to consistent candidate-level protocol application as a central challenge. Appendix~\ref{app:criteria_formulation} gives the complete setting and results.

\subsection{Controlled Oracle Retrieval}

To separate evidence availability from later selection, the oracle diagnostics improve retrieval while keeping GPT-5.4, prompts, settings, and workflow fixed on all 86 instances. Every workflow sees at most 200 distinct articles, including across agent retrieval calls, and the GPT-Researcher oracle keeps this budget. We call the standard ranking Actual. GT-first places reference articles first and fills remaining positions from the standard ranking. One-pass RAG uses $K=200$, and each GPT-Researcher call uses $K=20$. GT-only gives one-pass RAG the complete reference set alone. Appendices~\ref{app:oracle_design} and~\ref{app:oracle_full} give the construction and full results.

\begin{table}[t]
\centering
\caption{Core GPT-5.4 oracle results and a fixed top-12 diagnostic on 86 test instances (\%). Actual uses the standard ranking; GT-first places reference articles first; GT-only supplies only the complete reference set. The fixed rule uses the first 12 articles as its final list.}
\label{tab:oracle_summary}
\small
\setlength{\tabcolsep}{4pt}
\begin{tabular}{@{}llrr@{}}
\toprule
\textbf{System} & \textbf{Retrieval} & \textbf{Inc.R} & \textbf{Inc.P} \\
\midrule
\multirow{3}{*}{One-pass RAG} & Actual & 48.4 & 76.5 \\
 & GT-first & 48.6 & 77.0 \\
 & GT-only & 48.0 & 90.7 \\
\midrule
\multirow{2}{*}{GPT-Researcher} & Actual & 30.3 & 76.7 \\
 & GT-first & 48.8 & 88.4 \\
\midrule
\multirow{2}{*}{Fixed top-12 rule} & Actual & 42.3 & 39.7 \\
 & GT-first & 74.4 & 79.6 \\
\bottomrule
\end{tabular}
\end{table}

\textbf{Workflows respond differently to improved evidence availability.} For one-pass RAG, GT-first retrieval raises pool recall from 91.7\% to 100.0\%, while Inc.R changes only from 48.4\% to 48.6\%. GT-only input yields 48.0\% Inc.R. The reports have no cap on the number of listed articles. Eight completed reports select no included article and therefore receive zero precision; every article in the other 78 lists belongs to the reference set. GPT-Researcher responds more strongly: GT-first retrieval raises its pool recall from 72.3\% to 93.6\% and Inc.R from 30.3\% to 48.8\%. Better retrieval helps the iterative agent, and both workflows still omit reference articles in these controlled settings. Appendix~\ref{app:oracle_design} gives the oracle construction, and Appendix~\ref{app:oracle_full} reports the complete pool and retention results.

\textbf{A fixed top-12 rule exposes a large selection gap under oracle ranking.} As a non-LLM diagnostic, we treat the first 12 retrieved articles as the final list. This list size is within the 10.2 to 12.6 mean range of the three one-pass RAG models. On the standard MA-Retriever ranking, the rule reaches 42.3\% Inc.R and 39.7\% Inc.P, compared with 48.4\% and 76.5\% for GPT-5.4 RAG. Under reference-first ranking, the same rule reaches 74.4\% Inc.R and 79.6\% Inc.P, compared with 48.6\% and 77.0\% for reference-first RAG. This controlled comparison shows that final selection can remain incomplete even when reference articles are placed first. Appendix~\ref{app:oracle_cutoff} reports the same diagnostic at four list sizes.

\section{Related Work}
\label{sec:related}

\subsection{Meta-Analysis Automation}

Automation research covers the full meta-analysis workflow~\cite{tsafnat2014automation}, text mining for citation screening~\cite{omaraeves2015textmining}, and machine-learning tools for evidence synthesis~\cite{marshall2019automation}. ASReview uses active learning to reduce screening effort~\cite{vandeschoot2021asreview}, and the spiral approach combines machine learning with an iterative workflow~\cite{saeidmehr2024systematic}. Recent LLM studies still find that fully automatic use requires expert oversight~\cite{Khraisha_2024,lieberum2025large}.

CLEF TAR established retrieval and screening tasks in empirical medicine~\cite{kanoulas2017clef}. SYNERGY provides 169,288 records for 26 tasks~\cite{debruin2023synergy}, and CSMeD consolidates nine screening collections covering 325 tasks~\cite{kusa2023csmed}. TrialReviewBench covers retrieval, screening, and extraction for 100 clinical tasks using task-specific 2,000-citation pools~\cite{trialreviewbench2025}. MetaSyn instead uses one shared corpus for 422 instances and follows article IDs to the final evidence list.

\subsection{Retrieval and LLM Research Agents}

RAG supplies an LLM with retrieved evidence~\cite{lewis2020retrieval}, with designs surveyed by Gao et al.~\cite{gao2024rag}. Fixed retrieval can use BM25~\cite{bm25}, DPR~\cite{karpukhin2020dense}, or BGE~\cite{bge}. ReAct interleaves reasoning with tools~\cite{yao2023react}, and Zhu et al. survey LLM agents for information retrieval~\cite{zhu2025llmir}. Active RAG~\cite{jiang2023active} and DRAGIN~\cite{su2024dragin} retrieve adaptively; Search-R1~\cite{jin2025search} and WebThinker~\cite{li2025webthinker} develop search-based reasoning. Robust training can also reduce retrieval-error effects~\cite{tu2025robust}.

Graph-based~\cite{edge2024local} and multi-hop retrieval~\cite{mrrfv} change how evidence is gathered, and ranking quality can depend on the downstream model~\cite{salemi2024towards}. In evidence synthesis, topical relevance cannot enforce population, design, comparator, outcome, and other protocol rules. MetaSyn therefore reports retrieval and final selection separately.

\subsection{Benchmarks for Scientific Synthesis}

Multi-XScience studies multi-article summarization~\cite{lu2020multi}, alongside broader multi-document work~\cite{ma2022multi}. Scientific-survey systems include STORM~\cite{shao2024storm}, AutoSurvey~\cite{autosurvey}, the framework of Lai et al.~\cite{lai2024instruct}, SurveyForge~\cite{yan2025surveyforge}, and SurveyX~\cite{liang2025surveyx}.

Evaluation resources cover survey generation in NLPCC~\cite{nlpcc2024} and Gao et al.~\cite{gao2024evaluating}, scientific synthesis in SurGE~\cite{surge} and DeepScholar-Bench~\cite{deepscholar2025}, factual support in FActScore~\cite{min2023factscore} and VeriScore~\cite{song2024veriscore}, and long context in LongBench~\cite{bai2024longbench} and L-Eval~\cite{an2024eval}. MetaSyn adds a protocol-defined target and traces article IDs through retrieval, selection, and synthesis. Its 422 instances and shared corpus provide a larger fixed evaluation set than the related resources in Appendix~\ref{app:related_scope}.

\section{Limitations}
\label{sec:limitations}

MetaSyn evaluates retrieval, protocol-based selection, and synthesis in a controlled 140,585-record PubMed corpus. Each reference set covers the PubMed-linked part of a source review's list, so Inc.P measures agreement with that subset. The linked IDs support future study-level effect synthesis, which requires broader article access, cross-report linkage, and outcome harmonization.

Training and test contain distinct source reviews. Prompts omit source-review titles, IDs, and reports; retrieval removes the source review; and retriever training excludes test-linked positives. Some decisions require full text and expert assessment, while model pretraining remains unauditable.

\section{Conclusion}
\label{sec:conclusion}

MetaSyn is a dataset and benchmark for evaluating LLM agents on protocol-driven meta-analysis. It links 422 expert-curated meta-analyses, their research questions, protocols, published included-study lists, and synthesis labels to a shared 140,585-article PubMed corpus through stable IDs. Its stage-wise protocol evaluates retrieval, evidence selection, criteria reporting, and synthesis, making failures observable throughout the evaluated workflow. Baseline experiments demonstrate the benchmark's value: retrieval reaches 91.7\% R@200, while end-to-end inclusion recall peaks at 51.2\%, and oracle input still leaves substantial evidence unlisted. These results identify protocol-based selection as a distinct challenge and establish MetaSyn as a reproducible resource for developing broad, auditable, and complete evidence-synthesis systems.

\clearpage
\bibliographystyle{ACM-Reference-Format}
\bibliography{references}

\clearpage
\appendix
\small

\section{Detailed Terminology}
\label{app:terminology}

Table~\ref{tab:terminology_detailed} expands the compact terminology in Section~\ref{sec:terminology}. These distinctions matter because a source review reports a meta-analysis, while the primary studies it includes may be reported in one or more other articles. Selection is the overall process of choosing evidence, while screening is a candidate-level decision. The final report makes the included evidence observable through its explicit included-article list. Standard retrieval uses the original ranking. Reference-first retrieval places reference articles first, and reference-only input contains only the complete reference set. The result tables use the short labels Actual, GT-first, and GT-only.

\begin{table*}[t]
\centering
\caption{Detailed terminology used in MetaSyn. Article labels are defined separately for each benchmark instance.}
\label{tab:terminology_detailed}
\small
\renewcommand{\arraystretch}{1.05}
\begin{tabular}{@{}p{0.18\textwidth}p{0.45\textwidth}p{0.30\textwidth}@{}}
\toprule
\textbf{Term} & \textbf{Definition} & \textbf{Example} \\
\midrule
Source review & The expert-authored publication from which one benchmark instance derives its protocol, evidence list, and synthesis labels. & A Nature Portfolio article on maintenance antidepressant therapy. \\
Benchmark instance & One source review together with its structured question, reference set, protocol labels, and synthesis labels. & The antidepressant example and all MetaSyn fields derived from it. \\
Meta-analysis protocol & The eligibility logic, including PI/ECO fields, study design, date bounds, and stated inclusion/exclusion rules. & Patients in remission; continued antidepressant versus placebo; relapse during follow-up. \\
Included-study list & The source review's published set of included studies or study reports, recorded before PubMed matching. & The 40 trials reported by the source review. \\
Evidence synthesis & The conclusion and key findings drawn from the included evidence. & The source review's conclusion about relapse after treatment discontinuation. \\
Primary study & One investigation contributing evidence to a meta-analysis; several articles may report the same study. & One randomized maintenance trial. \\
Study report & A publication that reports a primary study; one study may have more than one report. & The journal article reporting a randomized maintenance trial. \\
Corpus article & A PubMed record in MetaSyn that represents a published article. & One record with a title, abstract, and optional PMC full text. \\
Candidate article & A corpus article considered for possible inclusion; eligibility is determined for each instance. & One result returned by BM25, BGE, or MA-Retriever. \\
Included article & A corpus record whose title matches one entry in the source review's included-study list. & A linked trial report from the source review's list. \\
Reference set & All included articles linked to one benchmark instance and used to score it. & The 33 linked articles for the antidepressant example. \\
Retrieved pool & The top-$K$ articles returned by fixed retrieval, or the deduplicated union returned by an agent. & A top-200 RAG context. \\
Retrieval query & The research question and PI/ECO text submitted once to a fixed retriever. & The query used to form a top-200 RAG pool. \\
Retrieval call & One retrieval request issued by an iterative agent; one benchmark run may contain many calls. & One of GPT-Researcher's 13 retrieval calls in a run. \\
Predicted included article & A corpus article explicitly listed or cited as included in the generated report. & An article in the report's included-article list. \\
Generated report & The system output containing its protocol description, evidence selection, synthesis, and explicit included-article list. & The final RAG, ProtoMA, or deep-research report evaluated by MetaSyn. \\
Key insight & A substantive finding emphasized by the source review, such as an effect, comparison, subgroup pattern, or limitation. & Maintenance treatment reduces relapse, with heterogeneity across follow-up windows. \\
Conclusion direction & A Positive, Negative, or Mixed label for the source review's main conclusion. & Mixed for conflicting, null, heterogeneous, descriptive, or otherwise non-directional conclusions. \\
\bottomrule
\end{tabular}
\end{table*}

\section{Additional Dataset Disclosure}
\label{app:data}

\subsection{Comparison with Related Resources}
\label{app:related_scope}

Table~\ref{tab:related_scope} compares evaluation scale and domain coverage, how each resource organizes candidate articles, and which stages it can evaluate. MetaSyn combines the largest fixed set of benchmark instances in this comparison with one shared corpus and article IDs that can be followed from retrieval to the final included-article list.

\begin{table}[H]
\centering
\caption{Scale and scope of closely related benchmark resources.}
\label{tab:related_scope}
\footnotesize
\setlength{\tabcolsep}{3pt}
\begin{tabular}{@{}p{0.23\columnwidth}p{0.19\columnwidth}p{0.50\columnwidth}@{}}
\toprule
\textbf{Resource} & \textbf{Task scale} & \textbf{Candidate collection and evaluation targets} \\
\midrule
CLEF TAR & Medical topics & Topic-specific retrieval and screening collections \\
SYNERGY & 26 tasks & 169,288 works in task-specific screening pools; citation decisions \\
CSMeD & 325 tasks & Nine citation-screening collections; separate CSMeD-FT full-text task \\
TrialReviewBench & 100 clinical tasks & 2,220 included studies; 2,000-citation screening pools, retrieval, and data extraction \\
MetaSyn & 422 instances, 11 domains & Shared 140,585-article corpus; retrieval, protocol-based selection, synthesis, and cross-stage article IDs \\
\bottomrule
\end{tabular}
\Description{A three-column table compares task scale, candidate organization, evaluation stages, and cross-stage article tracing for CLEF TAR, SYNERGY, CSMeD, TrialReviewBench, and MetaSyn.}
\end{table}

\subsection{Annotation Protocol and Provenance}
\label{app:annotation_provenance}

Approximately 50 annotators were selected from 76 applicants. Selection considered academic background, English-language reading proficiency, and availability because the task required sustained reading of scientific articles and supplements. The annotators were primarily senior undergraduate and graduate students from multiple universities. Every annotator received the same written protocol, task-specific instructions, and worked examples before annotation.

Candidate discovery used the literal keywords ``meta-analysis'' and ``systematic review'' in the \textit{Nature} Portfolio. The queries returned more than 60,000 hits; deduplication and scope filtering produced 34,375 candidates for manual screening. Annotators inspected open-access candidates and retained articles whose complete included-study lists could be reliably recovered from the article, its supplements, linked data repositories, or references reported with forest plots. The annotation protocol required complete titles or full citations for the included studies.

For each retained article, annotators recorded the research question, search metadata, eligibility criteria, included studies, and synthesis labels. They checked that each extracted list covered all included studies and compared its size with the count reported in the source review when available. Every retained record then underwent a second verification pass against the source review and its supplements, normally by another annotator. Uncertain records were corrected or removed, and records with unreliable evidence extraction or more than 300 included entries were outside the final benchmark scope. PubMed matching was performed only after the title lists had been verified. This process ties every released label to an inspectable record in the source review.

Search dates follow the source review's reporting. Annotators recorded the last reported search date or the stated endpoint of the publication period searched. A year-only endpoint was normalized to December 31, a month-only endpoint to that month's final day, and a complete date was retained as reported. The release stores the normalized date without a separate flag for its source type or original precision. The field is missing for 2 of 422 instances and 1 of 86 test instances, giving the 99.5\% and 98.8\% availability rates in Table~\ref{tab:data_stats}.

A small set of PI/ECO examples was structured manually. GLM-4.6 received the source review and drafted most remaining PI/ECO fields. Annotators corrected those drafts, and every retained record underwent a second verification pass. Verified titles were then matched to PubMed records to produce the released corpus IDs.

The shared corpus came mainly from PubMed results returned during title matching, together with a smaller set of randomly sampled PubMed records from the first-pass pool.

The release provides the verified titles and final linked IDs so that the matching statistics can be recomputed.

\subsection{Coverage and Dataset Fields}
\label{app:dataset_coverage}

The 422 meta-analysis instances are grouped into eleven domains based on source journal. Clinical specialties account for 67.3\% of the dataset; the rest covers digital health and AI in medicine, social sciences and humanities, environmental and planetary health, general biology, and multidisciplinary topics. We retained source reviews only when their included-study lists could be inspected directly.

\begin{figure*}[t]
\centering
\includegraphics[width=0.96\textwidth]{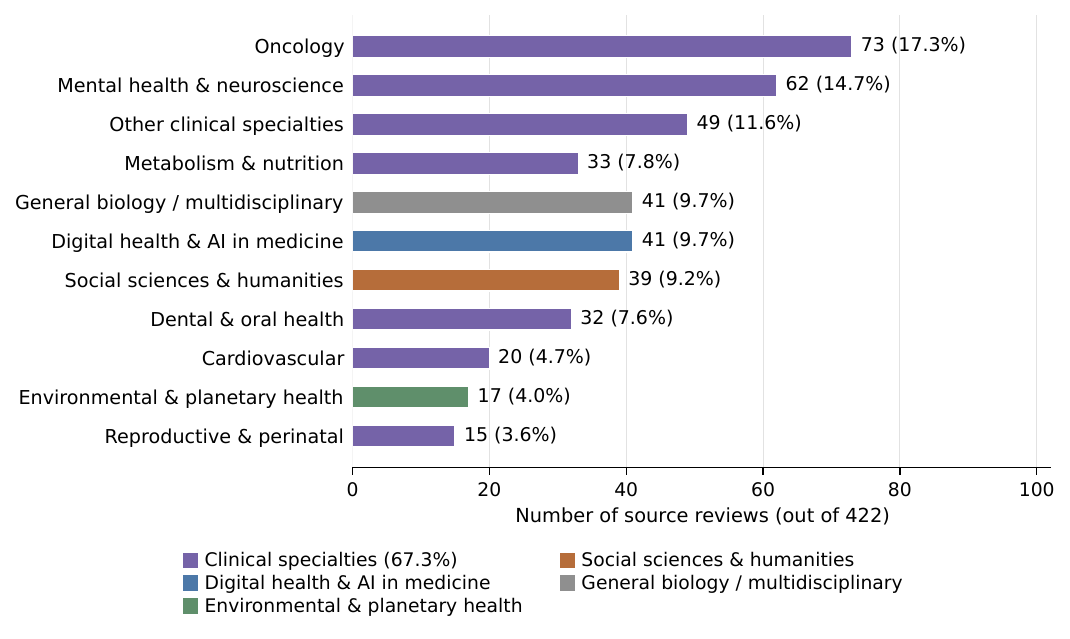}
\caption{Domain distribution of the 422 meta-analysis instances. Clinical specialties account for 67.3\%; the remaining 32.7\% spans digital health and AI in medicine, social sciences and humanities, environmental and planetary health, and general biology or multidisciplinary topics.}
\Description{Horizontal bars show the number of instances in each domain, with clinical specialties forming the majority and several non-clinical groups forming the remainder.}
\label{fig:fields_appendix}
\end{figure*}

Match rates vary across domains. Database coverage is one likely reason. Clinical meta-analyses, especially in oncology, often search EMBASE, trial registries, and conference proceedings as well as PubMed. Several environmental and social-science journals in MetaSyn have a larger share of PubMed-indexed included articles. The benchmark reports this variation because it determines how much of each instance can be evaluated in the shared corpus.

For each instance, the title-match rate is the number of linked PubMed articles divided by the number of nonempty titles transcribed from its published included-study list. We then average these instance-level rates equally.

MetaSyn also records synthesis fields beyond retrieval labels. An instance-level effect-size field is available for 174 instances and is unused by the main benchmark.

\subsection{Split and Training Scope}
\label{app:split_scope}

The released \texttt{test\_ids.json} defines an 86-instance test split with distinct source reviews; the other 336 instances form the training split. Instances with more included articles received higher sampling weight. We exclude every test-linked article from the positive retriever-training pairs, leaving 334 training instances that contribute pairs.

\subsection{Release, Versioning, and Responsible Use}
\label{app:release_scope}

\textbf{Construction records.} Annotators verified the included-study lists published by the source reviews; PubMed and corpus IDs were linked from the extracted titles. The release stores every verified title, including entries without a PubMed match, together with the deduplicated linked IDs. These fields support recomputation of the aggregate statistics and future matching through additional literature databases.

The public release provides the meta-analysis and corpus records as Hugging Face Parquet files, the 86 test-instance IDs, dataset and model cards, and retrieval, training, one-pass RAG, and evaluation code. Code and project-authored annotations use the MIT License. Identifiers, abstracts, source-review excerpts, and official PMC-derived sections retain their upstream terms and are not relicensed by MetaSyn.

\textbf{Versioning.} Each release keeps stable instance IDs, an immutable test split, and versioned corpus, retriever, and evaluation metadata. New instances can be added in later versions without changing the reported test set; results should state the dataset and protocol version they use.

\textbf{Scope and responsible use.} The reference set covers the PubMed-linked portion of each included-study list. Source reviews also report searches of EMBASE, CENTRAL, trial registries, conference proceedings, and other databases. Closely related corpus records remain plausible screening candidates, and some eligibility decisions require full text. Generation prompts omit the source-review title, ID, and report. Retrieval removes the source review before top-$K$ truncation, and retriever training excludes test-linked positive pairs. Public model pretraining remains outside our audit scope. Human oversight remains necessary in high-impact applications.

\newpage
\section{Supplementary Experiment Details}
\label{app:supplementary_experiments}

\subsection{Retrieval Behavior and Screening Cases}
\label{app:retrieval_cases}

\textbf{MA-Retriever gains most over BGE on instances with larger reference sets.} At $K=200$, MA-Retriever reaches perfect recall on 55 of 86 test instances, compared with 44 for BGE. At $K=100$, it improves 39 instances and performs worse on 7. Its mean R@100 gain grows from 3.1 points for instances with fewer than 10 linked articles to 9.7 points for instances with at least 50.

\textbf{High-ranked cases not selected by source reviews.} Instance 23 requires time-restricted eating together with exercise and an exercise-matched control. Its top-20 pool also contains protocols, secondary syntheses, and trials with different feeding or exercise conditions. Instance 194 requires colour-fundus diagnosis of pathological myopia; several high-ranked candidates use optical coherence tomography or target a different myopia label. Instance 28 focuses on engagement with synchronous remote weight-management interventions; its pool includes asynchronous, self-directed, and protocol articles. These explanations come from the recorded PI/ECO fields and candidate abstracts.

\subsection{Output and Citation Diagnostics}
\label{app:output_diagnostics}

We use list length and citation mapping to distinguish short evidence lists from failures to identify a retrieved corpus article. The following means are computed directly from the 86 task records. The prompts place no cap on final-list length. ``Unmapped'' citations are explicit report entries without a match to a retrieved corpus article and remain in the Inc.P denominator.

\begin{table}[H]
\centering
\caption{Mean final included-article list size for systems using MA-Retriever.}
\label{tab:listing_budget}
\small
\begin{tabular}{@{}lrrr@{}}
\toprule
\textbf{System} & \textbf{Mean list size} & \textbf{Unmapped} & \textbf{Inc.R (\%)} \\
\midrule
RAG DeepSeek-V4-Pro & 12.6 & 0.01 & 51.2 \\
RAG GLM-5.1 & 12.0 & 0.01 & 51.0 \\
RAG GPT-5.4 & 10.2 & 0.13 & 48.4 \\
ProtoMA GPT-5.4 & 9.3 & 0.0 & 39.3 \\
GPT-Researcher & 5.5 & 0.0 & 30.3 \\
OpenDR & 4.5 & 0.0 & 27.2 \\
\bottomrule
\end{tabular}
\end{table}

Table~\ref{tab:mapping_diagnostic} reports mapped-only precision, macro-averaged over instances with at least one mapped citation. Unmapped citations are removed only for this diagnostic; reported Inc.P retains them.

\begin{table*}[t]
\centering
\caption{Citation mapping diagnostic for MA-Retriever RAG. Correct is the mean number of predicted corpus articles in the instance's reference set. Mapped is the mean number of predicted entries matched to a retrieved corpus article. Unmapped is the mean number of predicted entries without such a match.}
\label{tab:mapping_diagnostic}
\small
\begin{tabular*}{\textwidth}{@{\extracolsep{\fill}}lrrrr@{}}
\toprule
\textbf{RAG model} & \textbf{Correct} & \textbf{Mapped} & \textbf{Unmapped} & \textbf{Mapped-only P (\%)} \\
\midrule
DeepSeek-V4-Pro & 9.44 & 12.56 & 0.01 & 78.4 \\
GLM-5.1 & 9.51 & 11.98 & 0.01 & 82.2 \\
GPT-5.4 & 8.52 & 10.08 & 0.13 & 86.9 \\
\bottomrule
\end{tabular*}
\end{table*}

Unmapped entries average only 0.01 to 0.13 per RAG report. The observed Inc.P differences therefore arise mainly from which mapped corpus articles are selected, while list length explains part of the recall difference across workflows.

\subsection{Pipeline Traces and Cross-System Union}
\label{app:pipeline_traces}

\textbf{ProtoMA stage loss.} The macro decomposition assigns 20.7\% of the reference set to retrieval loss, 40.0\% to explicit screening loss, and 39.3\% to the final list. Its mean pool contains 72.35 articles, including 12.62 reference articles, of which 6.27 are finally listed. Its final predicted list contains 9.3 entries per instance on average. Every positive screening ID reaches the final ID list, as required by ProtoMA's recorded workflow.

\textbf{Union across configurations.} Taking the union of the twelve standard-retrieval configurations gives 95.0\% macro retrieval recall; 59 of 86 instances have complete union coverage, and 132 of the 1,677 test instance/article pairs remain unretrieved. The union of final article lists reaches 75.7\% macro recall, leaving a measurable gap even when multiple workflows are combined. These values summarize the twelve tested configurations.

\subsection{Date Filtering}
\label{app:date_filtering}

This supplementary setting measures the effect of applying known temporal bounds. We use the final saved MA-Retriever ranking, remove records published after the source review's stored search end year, and refill the top-20 pool from deeper ranks. Same-year and missing-year records remain; the one instance without an end year keeps its original pool. Across all 86 instances, R@20 rises from 53.5\% to 66.4\% and P@20 from 34.0\% to 43.9\%. Of the 1,135 returned records not selected by source reviews, 627 (55.2\%) are later-year records. Twelve records selected by source reviews also appear after the stored cutoff, consistent with differences among online, issue, and indexing dates. The filtered pool still contains many candidates that require protocol screening.

\subsection{Eligibility Criteria Formulation}
\label{app:criteria_formulation}

The main experiment supplies recorded criteria to hold the protocol fixed. To test the additional task faced by a more autonomous assistant, the self-formulated condition removes those criteria and asks the workflow to formulate its own. It retains the research question, PI/ECO, search dates, model, retriever, generation settings, and output format. Table~\ref{tab:self_formulated_criteria_appendix} reports the complete results.

\begin{table}[H]
\centering
\caption{Complete eligibility criteria input ablation on 86 test instances. Pool R is candidate-pool recall. SQ is on a 1 to 5 scale; all other metrics are macro-averaged percentages.}
\label{tab:self_formulated_criteria_appendix}
\normalsize
\setlength{\tabcolsep}{4.2pt}
\begin{tabular}{@{}lrrrr@{}}
\toprule
\textbf{Metric} & \textbf{RAG G} & \textbf{RAG S} & \textbf{GPR G} & \textbf{GPR S} \\
\midrule
Pool R & 91.7 & 91.7 & 72.3 & 71.7 \\
Inc.R & 48.4 & 47.4 & 30.3 & 35.8 \\
Inc.P & 76.5 & 78.5 & 76.7 & 71.4 \\
Inc.F1 & 55.8 & 55.2 & 40.3 & 43.9 \\
Scr.A & 95.1 & 94.8 & 83.6 & 83.8 \\
Inc.C & 56.8 & 52.6 & 61.1 & 51.2 \\
Exc.C & 47.4 & 47.3 & 57.7 & 46.9 \\
Dir.A & 55.8 & 57.0 & 52.3 & 60.5 \\
Insights & 34.7 & 35.3 & 35.8 & 35.6 \\
SQ & 2.65 & 3.21 & 4.07 & 3.99 \\
\bottomrule
\end{tabular}
\vspace{0.25em}
\begin{flushleft}
\footnotesize G means recorded criteria are given; S means the workflow formulates them. GPR denotes GPT-Researcher.
\end{flushleft}
\end{table}

This ablation separates criteria reporting from evidence selection. One-pass RAG changes little in Inc.R when it formulates the criteria. GPT-Researcher gains 5.5 points in Inc.R and loses 5.3 points in Inc.P, while its criteria agreement declines. The main difficulty therefore lies in applying the protocol consistently to candidates.

\subsection{Descriptive System Profiles}
\label{app:system_profiles}

This figure provides a compact view of representative systems across the metrics in Table~\ref{tab:pipeline_results}. Scr.A changes partly because the pool contains many more articles not selected by the source review than selected articles. We therefore read it alongside Inc.R, Inc.P, and Inc.F1.

\begin{figure*}[t]
\centering
\begin{minipage}[t]{0.48\textwidth}
\centering
\textbf{(a) System profiles}\par\smallskip
\includegraphics[width=\linewidth]{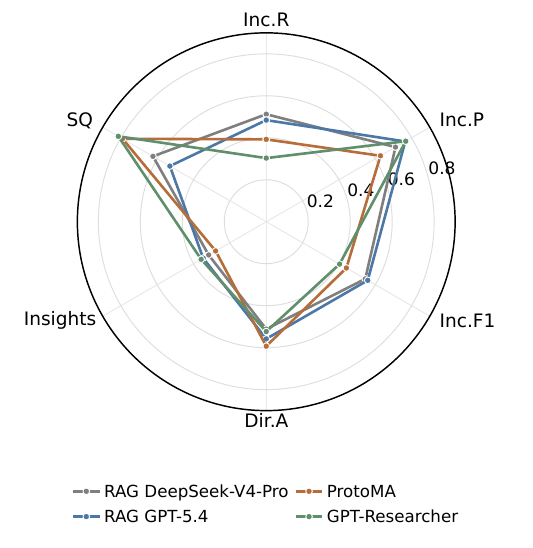}
\end{minipage}\hfill
\begin{minipage}[t]{0.48\textwidth}
\centering
\textbf{(b) Conclusion direction}\par\smallskip
\includegraphics[width=\linewidth]{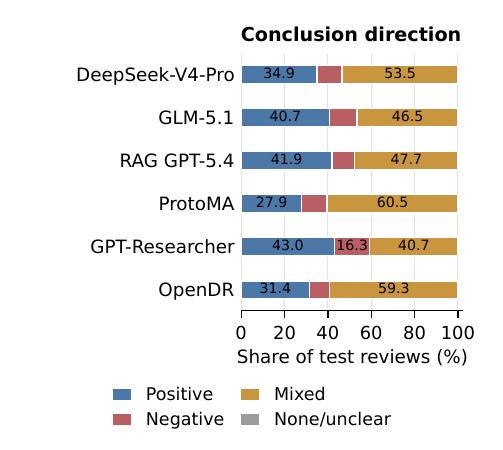}
\end{minipage}
\caption{Representative pipeline profiles. Panel (a) compares six selected metrics normalized to $[0,1]$; SQ is divided by 5. Panel (b) shows evaluator-extracted conclusion directions. ``Unclear'' means the report cannot be mapped to Positive, Negative, or Mixed. Table~\ref{tab:pipeline_results} provides all nine exact values.}
\Description{Two panels compare representative systems across six normalized metrics and show their extracted Positive, Negative, Mixed, and unclear conclusion directions.}
\label{fig:profiles_appendix}
\end{figure*}

The profiles show the same recall and precision regimes as the exact table: fixed-pool RAG favors broader evidence lists, while the deep-research agents favor shorter lists with higher precision. Conclusion direction varies separately from evidence coverage.

\section{Token Use Details}
\label{app:token}

\subsection{Measurement and Subset}
\label{app:token_measurement}

We count input and output tokens across all model calls and separately count tokens in returned titles and abstracts. The depth analysis divides instances into four strata by reference-set size and selects three instances from each stratum. One-pass RAG uses $K\in\{5,10,20,50,100,200\}$, and GPT-Researcher uses $K\in\{5,10,20\}$ for every retrieval call. We also record retrieval calls, returned article instances, unique articles, repeated articles, and elapsed time. The released subset file records the 12 task IDs and selection statistics.

\subsection{Depth and Retrieval Behavior}
\label{app:token_results}

\begin{figure*}[t]
\centering
\includegraphics[width=0.96\textwidth,trim=0 17pt 0 0,clip]{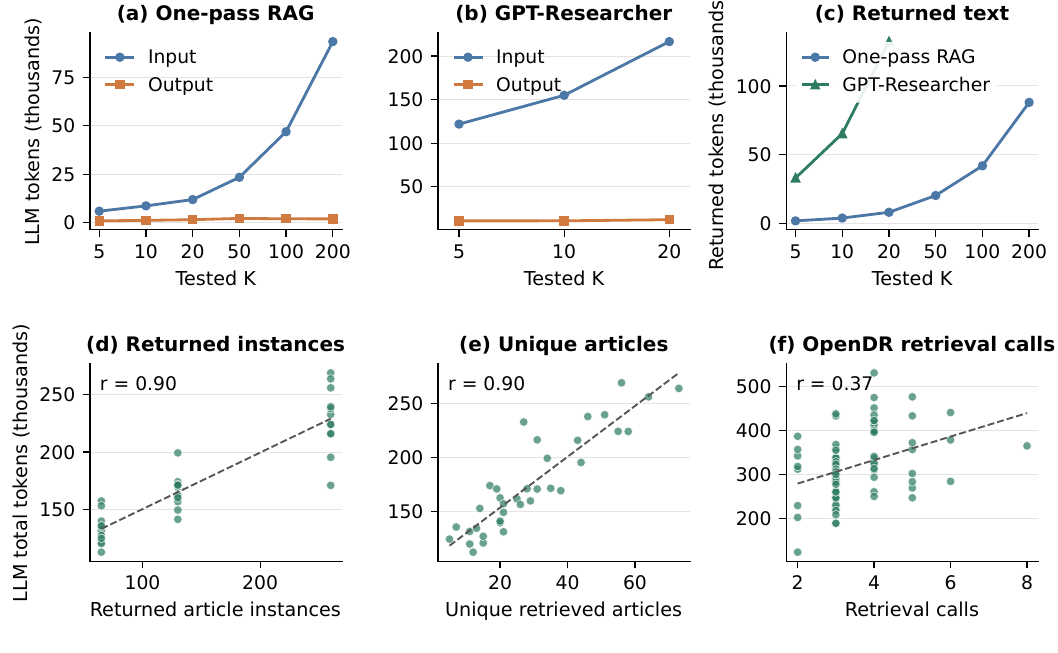}
\caption{Token use with MA-Retriever. Panels (a) and (b) show LLM input and output tokens, and panel (c) shows tokens in returned titles and abstracts at the tested $K$ values on the same 12-instance subset. Panels (d) and (e) relate GPT-Researcher's total LLM tokens to all returned article instances and unique articles. Panel (f) relates OpenDR's total LLM tokens to retrieval calls across all 86 instances at fixed $K=20$.}
\Description{Six panels compare token growth for one-pass RAG and GPT-Researcher, returned article instances, unique articles, and OpenDR retrieval calls.}
\label{fig:token_pressure}
\end{figure*}

Figure~\ref{fig:token_pressure} reports model-token usage and retrieval behavior for this subset. The main text gives the corresponding mean changes in retrieval recall and final inclusion recall.

One-pass RAG grows mainly with the returned title-and-abstract text. GPT-Researcher also repeats retrieval across calls: increasing $K$ from 5 to 20 raises returned article instances from 65 to 260, while unique articles rise from 13.8 to 48.2. OpenDR keeps $K=20$ across all 86 instances, allowing the number of retrieval calls to vary with the agent's behavior. Its total model tokens increase with retrieval calls ($r=0.37$, $p<0.001$).

\section{Controlled Oracle Retrieval}
\label{app:oracle_section}

\subsection{Oracle Construction}
\label{app:oracle_design}

We use two oracle interventions. Reference-first retrieval places reference articles before standard results while preserving the requested pool size. Reference-only RAG supplies the complete reference set and no other candidates.

Both one-pass RAG and GPT-Researcher use GPT-5.4 and receive the full recorded protocol for downstream selection. The one-pass retriever uses the fixed PI/ECO query, while GPT-Researcher generates queries from its full task prompt. For each request, the reference-first oracle ranks the instance's reference articles with MA-Retriever. If there are at least $K$, it returns the top $K$ reference articles for that request. Otherwise, it returns all reference articles first and fills the remaining positions from the standard ranking. One-pass RAG uses $K=200$; GPT-Researcher uses $K=20$ per retrieval call. Reference-only RAG uses the same prompt and generation settings but bypasses ranking and filler candidates.

\subsection{Full Oracle Results}
\label{app:oracle_full}

Table~\ref{tab:oracle_results} gives the full results for the reference-first and reference-only interventions defined above. Let $G$ be the linked reference set, $P$ the deduplicated candidate pool exposed to the workflow, and $L$ the deduplicated final list, with matched entries represented by corpus IDs. Pool Recall asks what fraction of $G$ appears in $P$. Inclusion Recall asks what fraction of $G$ appears in $L$. Retention asks what fraction of the retrieved reference articles, $P\cap G$, also appears in $L$. The retrieved-but-unlisted rate uses all of $G$ as its denominator and counts reference articles that appear in $P$ but not in $L$.

\begin{table}[H]
\centering
\caption{Controlled retrieval diagnostics on 86 test instances (\%). Pool R is reference coverage in the exposed candidate pool. Retention is the share of retrieved reference articles that reach the final list. Unlisted is the share of the full reference set that is retrieved but absent from the final list.}
\label{tab:oracle_results}
\small
\setlength{\tabcolsep}{3.1pt}
\begin{tabular}{@{}lrrrrr@{}}
\toprule
& \multicolumn{3}{c}{\textbf{One-pass RAG}} & \multicolumn{2}{c}{\textbf{GPT-Researcher}} \\
\cmidrule(lr){2-4}\cmidrule(l){5-6}
\textbf{Metric} & \textbf{Actual} & \textbf{GT-first} & \textbf{GT-only} & \textbf{Actual} & \textbf{GT-first} \\
\midrule
Pool R & 91.7 & 100.0 & 100.0 & 72.3 & 93.6 \\
Inc.R & 48.4 & 48.6 & 48.0 & 30.3 & 48.8 \\
Inc.P & 76.5 & 77.0 & 90.7 & 76.7 & 88.4 \\
Inc.F1 & 55.8 & 55.7 & 59.7 & 40.3 & 59.1 \\
Retention & 51.0 & 48.6 & 48.0 & 40.8 & 51.4 \\
Unlisted & 43.3 & 51.4 & 52.0 & 42.0 & 44.7 \\
\bottomrule
\end{tabular}
\Description{A transposed table compares actual, reference-first, and reference-only candidate pools for one-pass RAG and GPT-Researcher. It reports pool coverage, final-list recall and precision, retention, and retrieved reference articles omitted from the final list.}
\end{table}

\subsection{Fixed Rank-Cutoff Diagnostic}
\label{app:oracle_cutoff}

A fixed rule predicts the first $N$ articles as its final list. On the standard MA-Retriever ranking, Inc.R/Inc.P are 24.2/48.1\% at $N=5$, 36.9/43.2\% at $N=9$, 42.3/39.7\% at $N=12$, and 52.8/34.9\% at $N=19$. On the reference-first ranking, the same rule reaches 46.2/94.7\%, 66.3/87.0\%, 74.4/79.6\%, and 85.3/66.0\%, respectively. These reference-first values form a label-informed diagnostic.

\section{Metric Validation}
\label{app:metric_validation}

\subsection{Validation Protocol}
\label{app:validation_protocol}

Four metrics, Inc.R, Inc.P, Inc.F1, and Scr.A, use corpus IDs; unmatched predicted citations remain in the Inc.P denominator and receive no reference match. The other five are automated text metrics based on embedding similarity, semantic comparison, or direction extraction. We test whether their pairwise preferences agree with human judgments.

We sampled 30 test instances with higher weight for instances containing more included articles and built three representative comparisons spanning retriever and workflow changes. The eight annotators were selected from the dataset annotation team and were senior undergraduate, master's, or doctoral students. They received task instructions and examples. Each task showed the reference answer and two anonymous system outputs; system identities and the purpose of each comparison were hidden. All annotators saw the tasks in the same order, while the A/B positions were randomized. Annotators judged inclusion criteria, exclusion criteria, conclusion direction, and key insights for every pair, plus report structure for one pair. This produced 90 paired-output comparisons and 390 dimension-level judgments. All eight annotators rated each judgment independently on a five-point scale.

For agreement, strongly A and weakly A become A, the middle rating remains a tie, and the two B ratings become B. The automatic label is the sign of the score difference between outputs. For conclusion direction, tie against tie counts as agreement because both reports either match or miss the reference. Ties are removed for the other dimensions. We also compute Spearman $\rho$ between the mean five-point human rating and the automatic score difference.

Evaluator to human agreement can exceed average human pair agreement because it compares each annotator with one deterministic label. Spearman correlation is the main measure of whether metric differences track human preferences.

\subsection{Validation Results}
\label{app:validation_results}

Direction has 93.5\% human pair agreement and 91.2\% evaluator to human agreement; its rank correlation is $\rho=+0.82$. Human pair agreement is about 76\% for both criteria dimensions. The evaluator agrees more often on exclusion criteria (68.1\%, $\rho=+0.59$) than inclusion criteria (61.3\%, $\rho=+0.24$). Key-insight and structure scores also correlate positively but less strongly. Dir.A has the strongest support for close comparisons of the tested metrics, but it covers only the three-way conclusion label. Exc.C and Insights support system-level comparisons; Inc.C and SQ are descriptive profile measures.

\clearpage
\begin{figure}[H]
\centering
\includegraphics[width=\columnwidth]{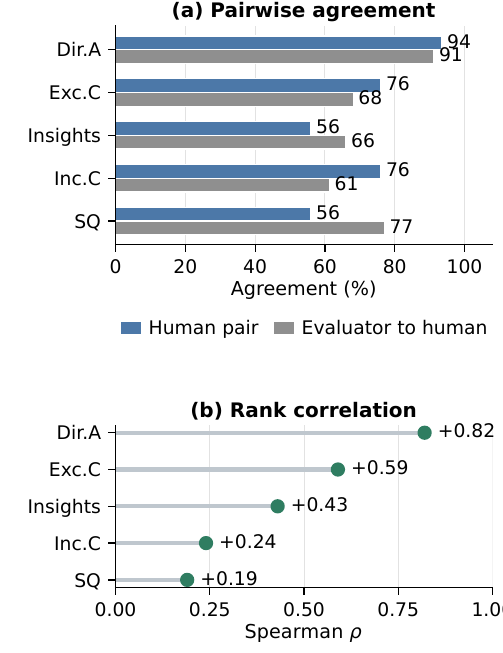}
\caption{Human validation of the evaluator-dependent metrics. The upper panel compares human-human pairwise agreement with evaluator-human agreement. The lower panel reports Spearman correlation between metric differences and mean human preferences. Direction is the strongest fine-grained signal; exclusion criteria and insights support broader comparisons.}
\Description{Two aligned panels report five metrics. The upper horizontal grouped bars show human pair agreement and evaluator-to-human agreement in percent. The lower horizontal dot plot shows Spearman rank correlation for the same metrics.}
\label{fig:annotation_appendix}
\end{figure}

\section{Reproducibility Details}
\label{app:reproducibility}

\subsection{Retriever Training and Evaluation}
\label{app:retrieval_setup}
Retrieval queries use the same structured format during training and evaluation: research question, Population, Intervention or Exposure, Comparison, and Outcome. BM25 uses $k_1=1.5$ and $b=0.75$. Dense BGE and MA-Retriever share the BGE-large-en-v1.5 architecture and use the same FAISS indexing method. MA-Retriever uses a 40-instance training-side validation set. We select the epoch with the highest R@20, using R@100 and R@200 as successive tie-breakers. We select epoch 1 and retrain the model for one epoch on the training split, producing 5,585 pairs from 334 of the 336 instances after excluding positive pairs whose articles are linked to a test instance.

\subsection{Generation and Workflow Settings}
\label{app:generation_setup}

Generation uses DeepSeek-V4-Pro, GLM-5.1, and GPT-5.4 as specified in the main results; automated evaluation uses GPT-5.5. Each run records the exact model name, settings, and input/output token counts.

The one-pass fixed-pool experiments use 200 candidates and provide each title, publication year, and abstract in rank order in one prompt. The main one-pass RAG and agent conditions also provide the recorded search dates and eligibility criteria. The self-formulated ablation removes those criteria and asks the workflow to formulate them in the same report.

ProtoMA formulates its criteria from PI/ECO and the recorded date bounds. It generates three to five MA-Retriever queries with three query-formulation examples from the training instances. It merges and deduplicates results under a cap of 200, screens in batches of 25, truncates abstracts to 500 characters during screening, and uses title plus abstract capped at 3,000 characters for extraction.

Agent retrieval calls use MA-Retriever with $K=20$. OpenDR uses at most two concurrent research units, two researcher iterations, ten tool calls per researcher, ten retrieval calls, and 100 fetches. A fetch can read only an article returned by an earlier retrieval call and supplies at most 3,000 characters of its available corpus sections.

\subsection{Automated Evaluator}
\label{app:evaluator_setup}

GPT-5.5 extracts criteria, conclusion direction, and key insights from the generated reports. Inc.C and Exc.C use the \nolinkurl{paraphrase-multilingual-mpnet-base-v2} checkpoint with normalized cosine similarity and soft-F1 after criteria extraction. Dir.A compares the extracted direction with the reference label; Insights and SQ use GPT-5.5 judgments. All evaluator calls use temperature zero. Report metrics are macro-averaged over the 86 instances; unmatched citations remain predicted inclusions in the Inc.P denominator and receive no reference match.

\end{document}